\documentclass[letterpaper]{article} % DO NOT CHANGE THIS
\usepackage[preprint]{aaai2027}  % DO NOT CHANGE THIS
\usepackage[hyphens]{url}  % DO NOT CHANGE THIS
\usepackage{graphicx} % DO NOT CHANGE THIS
\usepackage{natbib}  % DO NOT CHANGE THIS AND DO NOT ADD ANY OPTIONS TO IT
\usepackage{caption} % DO NOT CHANGE THIS AND DO NOT ADD ANY OPTIONS TO IT
\usepackage{algorithm}
\usepackage{algorithmic}

\usepackage{newfloat}
\usepackage{listings}
\DeclareCaptionStyle{ruled}{labelfont=normalfont,labelsep=colon,strut=off} % DO NOT CHANGE THIS
\floatstyle{ruled}
\newfloat{listing}{tb}{lst}{}
\floatname{listing}{Listing}
\usepackage{amsmath}
\usepackage{amssymb}  
\usepackage{amsfonts} 
\usepackage{multirow}
\usepackage{booktabs} % 
\usepackage{array}
\usepackage{xcolor}
\usepackage{tabularx}
\usepackage{makecell}
\usepackage{enumitem}

\usepackage{booktabs}

\title{CORE: Common Outcome Regularities from Action-Free Visual Demonstrations \\ for Robot Manipulation}
\author{
    \textsuperscript{\rm 1}Juyi Sheng,
    \textsuperscript{\rm 1}Mingxin Tan,
    \textsuperscript{\rm 1}Jincheng Li, 
    \textsuperscript{\rm 1}Mengyuan Liu
}
\affiliations{
    \textsuperscript{\rm 1}Peking University, Shenzhen Graduate School, Shenzhen, China\\
}

\begin{document}

\maketitle

\begin{abstract}
Robot imitation learning often relies on costly robot demonstrations, while abundant action-free visual demonstrations, such as human videos, are difficult to use because they lack robot-executable actions and suffer from embodiment gaps. We propose CORE, a policy learning framework that extracts Common Outcome Regularities (CORE) from visual demonstrations. Rather than transferring explicit actions across embodiments, CORE exploits a key observation: although successful trajectories for the same task can be diverse, their terminal states often share stable object configurations, spatial relations, and contact constraints. CORE first trains a terminal outcome encoder with contrastive and auxiliary temporal objectives, then aggregates successful terminal embeddings into visual goal prototypes, and finally injects these prototypes as global goal conditions into robot policies. Compared with language instructions, visual goal prototypes provide more concrete geometric and physical constraints for task completion. Across Meta-World, RoboTwin 2.0, and real-world manipulation, CORE improves the average success rate of the corresponding policy backbones by up to +3.9, +11.1, and +17.0 percentage points, respectively, and outperforms text-conditioned variants under the evaluated settings. The project and code are available at \url{https://logssim.github.io/CORE.github.io/}.
\end{abstract}

\begin{figure}[t]
    \centering
    \includegraphics[width=1.0\linewidth]{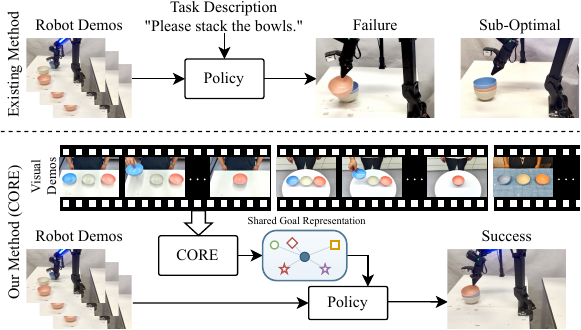}
   \caption{
    Outcome-guided policy learning with CORE. Task descriptions convey high-level
    intent but may under-specify the terminal geometry and contact constraints,
    leading to failure or suboptimal execution. CORE distills successful terminal
    observations from action-free visual demonstrations into a shared task-level
    goal representation that complements robot demonstrations and explicitly
    conditions the policy on the desired outcome.
    }
    \label{fig:motivation}
\end{figure}

\section{Introduction}

Imitation learning has substantially improved robot manipulation, but most end-to-end policies still rely on large amounts of robot-specific demonstrations, such as teleoperated or kinesthetic trajectories. Such data are expensive to collect and limit scalability across tasks, objects, and environments. In contrast, action-free visual demonstrations, such as human videos from the Internet and daily activities, are abundant and contain rich task priors, object interaction patterns, and physical common sense. Recent work has shown that such action-free videos can provide useful visual representations or reward signals for downstream robot control~\cite{ma2022vip}. However, directly using such data for robot policy learning remains challenging because they usually lack robot-executable action labels, and demonstrators differ substantially from the robot in morphology, action spaces, and control interfaces. Therefore, the central challenge is not simply to add visual demonstrations to the training set, but to extract supervision from them that is usable by robot policies despite the embodiment gap.

Beyond data availability, robot policies also require a precise specification of task goals. As shown in Fig~\ref{fig:motivation}, many methods use natural language instructions as task conditions because language is readily available and can express high-level task semantics, such as ``stack the bowls'' or ``place the object into the container.'' However, for contact-rich manipulation, language often describes only the task intent and leaves important physical details underspecified. It may not precisely specify object geometry, spatial poses, contact relations, or the terminal structure required for success. As a result, policies conditioned only on language may fail to focus on the outcome constraints that truly determine task completion, leading to sub-optimal or failed executions.

Motivated by this contrast, we revisit visual demonstrations from an outcome-level perspective. Our key observation is that motion trajectories for the same task can be diverse across different demonstrators, while their successful terminal states often share stable structures. For example, in a bowl-stacking task, different demonstrators may use different grasps, motion paths, and intermediate poses, yet successful executions consistently end in a stable nested configuration. Similarly, for placing, drawer manipulation, or stacking tasks, the process may vary, but the successful outcome satisfies consistent geometric relations, contact relations, and physical constraints. We refer to shared structures across successful terminal states as Common Outcome Regularities, which transfer across embodiments better than explicit actions and specify physical goals more concretely than language.

Based on this observation, we propose CORE, a framework that extracts Common Outcome Regularities from offline visual demonstrations and uses them for robot policy learning. CORE does not attempt to recover or transfer explicit actions, nor does it rely on a single goal image as the task condition. Instead, it learns visual outcome representations from successful terminal states and aggregates multiple successful terminals into a task-level visual goal prototype. This design changes the role of visual demonstrations from specifying \emph{how to move} to specifying \emph{what outcome counts as success}. The contrast in Fig.~\ref{fig:motivation} corresponds to a shift in goal representation: from under-specified language to outcome-level goal conditions induced from successful observations.

Concretely, CORE consists of three stages. First, it trains a terminal outcome encoder with terminal contrastive learning and auxiliary temporal objectives, encouraging the representation to focus on shared terminal structures rather than trajectory-specific motion details. Second, it aggregates embeddings of successful terminal states into visual goal prototypes, yielding stable task-level goal conditions. Third, it injects the goal prototype together with the current outcome embedding into a robot policy backbone, guiding action generation toward the desired terminal state. This goal-conditioning module is backbone-agnostic and can be combined with diffusion-based, flow-based, or other policies. Our contributions are summarized as follows:

\begin{itemize}
    \item We formulate visual-demonstration learning as the extraction of Common Outcome Regularities (CORE). Rather than struggling to map explicit actions across different embodiments, we extract shared structures directly from successful terminal states.
    \item We propose CORE, a three-stage framework that converts action-free visual demonstrations into robust visual goal conditions through terminal representation learning and prototype construction. Its backbone-agnostic design allows seamless injection into diverse downstream policies.
    \item Evaluations on Meta-World, RoboTwin 2.0, and real-world tasks show CORE consistently outperforms text-conditioned baselines, boosting underlying policy success rates by up to 3.9\%, 11.1\%, and 17.0\%, respectively.
\end{itemize}

\begin{figure*}[t]
    \centering
    \includegraphics[width=1.0\linewidth]{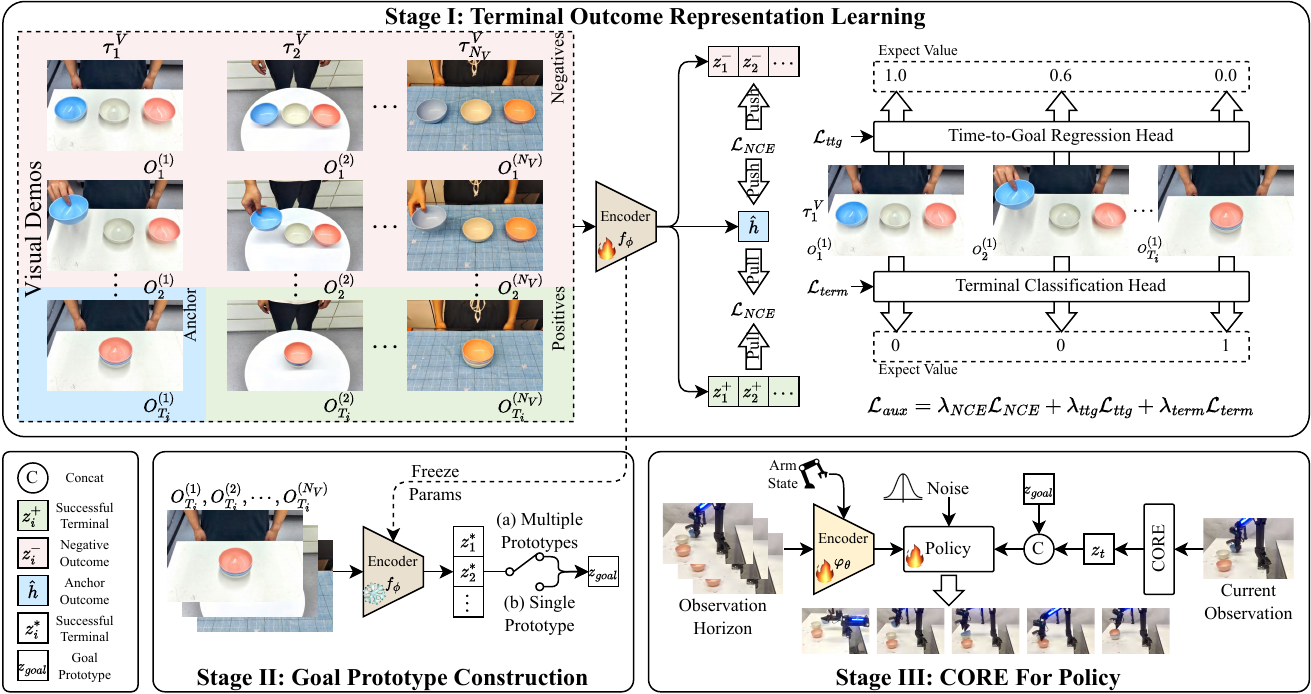}
   \caption{
    Overview of CORE. Stage I trains a terminal outcome encoder from action-free visual demonstrations. Stage II aggregates successful terminal embeddings into a visual goal prototype. Stage III combines the current outcome embedding with the goal prototype and injects the resulting goal feature into a policy backbone for outcome-guided action generation. 
    }
    \label{fig:framework}
\end{figure*}

\section{Related Work}

\paragraph{Visuomotor Policies and Vision-Language-Action Learning.}
Imitation learning for robot manipulation maps visual observations and robot states to executable actions. Recent action-chunking and generative policies improve temporal consistency and multimodal action modeling, including Diffusion Policy~\cite{chi2025diffusion}, DP3~\cite{Ze2024DP3}, FlowPolicy~\cite{zhang2025flowpolicy}, and MP1~\cite{sheng2026mp1}. Vision-language-action models further use natural language to specify task goals and improve semantic generalization~\cite{brohan2022rt,zitkovich2023rt,black2024pi_0,liu2025rdt}. However, these methods still mainly rely on robot demonstrations or large-scale robot data for action supervision, while language instructions may under-specify the geometry, spatial pose, contact relation, and terminal constraints required by contact-rich manipulation. CORE is complementary to these policy backbones: it does not replace the action predictor, but provides outcome-level visual goal conditions extracted from successful visual demonstrations.

\paragraph{Learning from Human Videos.}
Human videos provide scalable, rich priors for object interactions and physical constraints~\cite{lum2025crossing,chen2025vidbot,feng2026human}. However, their lack of explicit action labels and significant embodiment gaps hinder direct application to robot policies~\cite{lum2025crossing,kim2025uniskill,lepert2025masquerade}. Existing methods utilize these videos through latent action learning~\cite{ye2025latent,bi2026h,ye2026world,li2026causal}, trajectory recovery~\cite{chen2025vidbot,lum2025crossing}, video editing~\cite{lepert2025masquerade}, cross-embodiment skill representations~\cite{kim2025uniskill,lepert2025phantom}, or implicit visual rewards for reinforcement learning~\cite{ma2022vip}. While effective, these approaches typically require complex correspondences across actions, trajectories, or embodiments. Taking an outcome-level perspective, CORE avoids action recovery and implicit rewards. Instead, it extracts Common Outcome Regularities from successful demonstrations and explicitly aggregates them into task visual goal prototypes.

\begin{figure*}[t]
    \centering
    \includegraphics[width=0.8\linewidth]{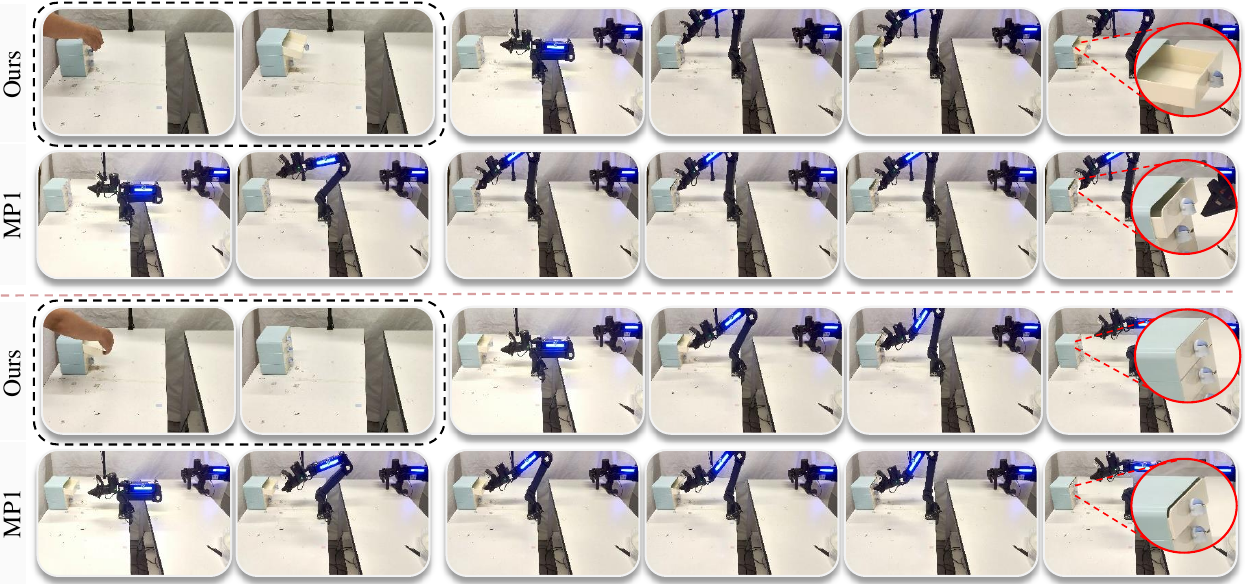}
    \caption{Qualitative results of CORE versus MP1 on the Open Drawer and Close Drawer tasks. MP1 fails to fully reach the target, while CORE leverages Common Outcome Regularities to complete the tasks successfully.}
    \label{fig:real_qualitative}
\end{figure*}

\section{Method}

\subsection{Overview}

CORE converts action-free visual demonstrations into visual goal conditions for robot policies. Let $\mathcal{D}_V=\{\tau_i^V\}_{i=1}^{N_V}$ denote the visual demonstration dataset, where each trajectory contains only goal-relevant observations, $\tau_i^V=(O_1^{(i)},\ldots,O_{T_i}^{(i)})$, and no explicit action labels are used. We assume demonstration-level success labels. For each successful trajectory, the last $w$ observations are treated as terminal positives; negatives are sampled from non-terminal observations and terminal observations from other tasks or failed trajectories.
CORE operates in three stages. Stage I trains a terminal outcome encoder $f_\phi$ to capture shared structures of successful terminal states. Stage II aggregates successful terminal embeddings into a task-level visual goal prototype $z_{\mathrm{goal}}$. Stage III concatenates the current outcome embedding $z_t$ with the goal prototype as $g_t=[z_t;z_{\mathrm{goal}}]$ and injects it into a policy backbone $B_\theta$ for action-chunk prediction.

\subsection{Stage I: Terminal Outcome Representation Learning}

To capture common outcome regularities, $f_\phi$ maps observations into an
$\ell_2$-normalized outcome space. For each contrastive sample, an anchor
observation and a positive observation are drawn from successful terminal
windows of the same task, with their embeddings denoted by $\hat{h}$ and
$z_i^+$, respectively. Negative embeddings $\{z_j^-\}$ are obtained from
non-terminal observations, terminal observations of failed demonstrations,
or successful terminal observations of other tasks. If failed demonstrations are unavailable, negatives are strictly drawn from the remaining two sources.

For a query $u$ and its positive $v$, we define the terminal contrastive
loss as
\begin{equation}
\ell_{\mathrm{NCE}}(u,v)
=
-\log
\frac{\exp(u^\top v/\tau_c)}
{\exp(u^\top v/\tau_c)
+\sum_{j=1}^{N_-}\exp(u^\top z_j^-/\tau_c)},
\label{eq:terminal_infonce}
\end{equation}
where $\tau_c$ is the contrastive temperature. We use a symmetric objective:
\begin{equation}
\mathcal{L}_{NCE}
=
\frac{1}{2}
\left[
\ell_{\mathrm{NCE}}(\hat{h},z_i^+)
+
\ell_{\mathrm{NCE}}(z_i^+,\hat{h})
\right].
\label{eq:symmetric_terminal_nce}
\end{equation}
This objective pulls together different visual realizations of the same
successful outcome while separating non-terminal, failed, and
task-irrelevant outcomes.

Terminal contrastive learning mainly constrains the final outcome region and
does not explicitly organize intermediate trajectory states. We therefore
introduce a Time-to-Goal Regression Head and a Terminal Classification Head
on the features produced by $f_\phi$. Let $s_i\in\{0,1\}$ be the success
label of $\tau_i^V$. Their supervision targets are
\begin{equation}
y_{i,t}^{ttg}
=
\frac{T_i-t}{T_i-1},
\qquad
y_{i,t}^{term}
=
\mathbb{I}
\left[
s_i=1
\land
t\geq T_i-w+1
\right].
\label{eq:auxiliary_targets}
\end{equation}
The time-to-goal target is defined for successful demonstrations and
decreases from $1$ to $0$ along the trajectory. The corresponding head is
trained with a Smooth-$\ell_1$ loss $\mathcal{L}_{ttg}$. The terminal head
is trained with binary cross-entropy $\mathcal{L}_{term}$, assigning a
positive label only to observations within successful terminal windows.
The complete Stage-I objective is
\begin{equation}
\mathcal{L}_{aux}
=
\lambda_{NCE}\mathcal{L}_{NCE}
+
\lambda_{ttg}\mathcal{L}_{ttg}
+
\lambda_{term}\mathcal{L}_{term}.
\label{eq:stage1_objective}
\end{equation}
Here, $\mathcal{L}_{ttg}$ introduces continuous trajectory-progress
information, whereas $\mathcal{L}_{term}$ sharpens the distinction between
successful terminal outcomes and non-terminal or failed outcomes. The two
auxiliary heads are removed after Stage I.

\subsection{Stage II: Goal Prototype Construction}

After Stage I, we freeze $f_\phi$ and encode the successful terminal
observations of each task into $\{z_i^\ast\}_{i=1}^{N}$. We then apply
$K$-Means, rank the resulting clusters by cardinality, and select the
Top-$K$ dominant clusters. Let $S_K$ denote the selected cluster indices
and $c_k$ the center of cluster $k$. The selected centers can be retained
as multiple prototypes; in the single-prototype setting, the visual goal
prototype is defined as
\begin{equation}
    z_{\mathrm{goal}}
    =
    \frac{
    \frac{1}{|S_K|}\sum_{k\in S_K} c_k
    }{
    \left\|
    \frac{1}{|S_K|}\sum_{k\in S_K} c_k
    \right\|_2
    }.
    \label{eq:goal_proto}
\end{equation}
Aggregating the dominant clusters reduces the influence of rare or noisy
terminal observations while preserving the major successful outcome modes,
yielding a fixed task-level visual goal prototype.

\subsection{Stage III: CORE for Policy}

At each policy step, the frozen CORE module uses $f_\phi$ to encode the
current observation into the current outcome representation $z_t$. As
denoted by $C$ in Fig.~\ref{fig:framework}, it is concatenated with the
goal prototype:
\begin{equation}
    z_t=f_\phi(O_t),
    \qquad
    C(z_t,z_{\mathrm{goal}})
    =
    [z_t;z_{\mathrm{goal}}].
    \label{eq:goal_condition}
\end{equation}
Meanwhile, the observation horizon and arm state are encoded by
$\varphi_\theta$. The policy combines these features with the CORE goal
condition and, when required by the generative backbone, noise to predict
an action chunk. CORE augments only the conditioning input and retains the
original action-learning objective of the selected policy backbone, allowing
it to be integrated with different action-chunking policies.

\begin{table*}[t]
    \centering
    \scriptsize
    \resizebox{1.0\linewidth}{!}{
    \begin{tabular}{
        l|
        >{\centering\arraybackslash}p{1.8cm}|
        >{\centering\arraybackslash}p{1.0cm}|
        >{\centering\arraybackslash}p{1.7cm}
        >{\centering\arraybackslash}p{1.7cm}
        >{\centering\arraybackslash}p{1.7cm}
        >{\centering\arraybackslash}p{2.0cm}|
        >{\centering\arraybackslash}p{1.8cm}}
    \toprule
    Method & Pub. & NFE  & Easy (21) & Medium (9) & Hard (5) & Very Hard (4) & \textbf{Avg.}
    \\  \midrule
    
    DP & RSS'23& 10 & 50.7$\pm${6.1} & 11.0$\pm${2.5}& 5.3$\pm${2.5}& 22.0$\pm${5.0}& 32.8$\pm${5.3} \\
    VIP & ICLR'23 & - & 44.6$\pm$7.9 & 9.8$\pm$5.2 & 11.1$\pm${3.9} & 15.4$\pm$4.3 & 29.3$\pm$6.4\\
    DP3 & RSS'24& 10 & 91.0$\pm${1.0} & 78.9$\pm${4.3} & 52.5$\pm${1.5} & 74.3$\pm${4.4} & 81.5$\pm${2.2} \\
    
    FlowPolicy & AAAI'25  & 1  & 87.6$\pm${1.9} & 75.7$\pm${6.7} & 36.6$\pm${3.5} & 67.5$\pm${2.2} & 76.3$\pm${3.2} \\ 
    MP1  & AAAI'26 & 1  & 90.5$\pm${0.6} & 78.7$\pm${4.1} & 55.1$\pm${5.1} & 75.5$\pm${4.7} & 81.7$\pm${2.4}\\ 
    DP3+TEXT & RSS'24& 10 & 91.2$\pm${1.3} & 79.7$\pm${4.9} & 52.3$\pm${1.8} & 75.1$\pm${3.9} &  81.9$\pm$2.5 \\
    MP1+TEXT  & AAAI'26 & 1  & 91.3$\pm${1.1} & 78.8$\pm${4.4} & 49.6$\pm${4.7} & 75.0$\pm${5.1} & 81.4$\pm$2.7\\
    \midrule
    DP3+CORE & - & 10 & 92.7$\pm${1.1} & 81.4$\pm${4.3} & 56.8$\pm${3.3} & 77.6$\pm${4.8} & 84.0$\pm${2.5} \\
    MP1+CORE  & - & 1 & \textbf{93.1}$\pm${1.0} & \textbf{81.9}$\pm${4.7} & \textbf{64.0}$\pm${4.2} & \textbf{81.4}$\pm${6.7} & \textbf{85.6}$\pm${2.8}
    \\ \bottomrule
    \end{tabular}
    }
    \caption{Performance of different methods on the Meta-World benchmark.}
    \label{results}
\end{table*}

\begin{table*}[h]
\centering
\resizebox{\textwidth}{!}{%
\scriptsize
\begin{tabular}{l|cccccccccc|c}
\toprule
\textbf{Method} 
& \shortstack{\textbf{adjust}\\\textbf{bottle}}
& \shortstack{\textbf{Dump Bin}\\\textbf{Bigbin}}
& \shortstack{\textbf{beat block}\\\textbf{hammer}}
& \shortstack{\textbf{move can}\\\textbf{pot}}
& \shortstack{\textbf{Stack blocks}\\\textbf{two}}
& \shortstack{\textbf{Stack bowls}\\\textbf{two}}
& \shortstack{\textbf{Handover}\\\textbf{block}}
& \shortstack{\textbf{Open}\\\textbf{laptop}}
& \shortstack{\textbf{open}\\\textbf{microwave}}
& \shortstack{\textbf{Stack bowls}\\\textbf{three}}
& \textbf{Avg.} \\
\midrule

DP 
& 90.6 
& 50.8 
& 48.2 
& 38.2
& 6.0 
& 61.0
& 12.0
& 46.6 
& 8.2 
& 62.8 
& 42.4 \\

VIP 
& 83.6 
& 51.8 
& 33.2 
& 48.4
& 8.0 
& 66.6
& 24.6
& 38.0 
& 19.4 
& 55.8 
& 42.9 \\

FlowPolicy 
& 93.6 
& 79.4 
& 56.8 
& 48.2 
& 18.8 
& 69.0 
& 38.6 
& 71.6 
& 21.8 
& 49.6 
& 54.7 \\

DP3 
& 98.6 
& 85.0 
& 80.2 
& 84.0 
& 30.6 
& 86.4 
& 88.4 
& 73.4 
& 79.4
& 69.2 
& 77.5 \\ 

MP1 
& 96.6 
& 63.8 
& 83.4 
& 83.4 
& 31.0 
& 85.0 
& 56.8 
& 80.4 
& 40.2 
& 62.4 
& 68.3 \\

DP3+TEXT
& 96.6 
& 86.2 
& 79.8 
& 84.6 
& 30.2 
& 86.0 
& 88.8 
& 73.8 
& 78.4
& 69.6 
& 77.4 \\ 

MP1+TEXT
& 96.8 
& 72.2 
& 82.6 
& 83.8
& 31.4 
& 85.8 
& 55.6 
& 80.4 
& 44.8 
& 63.2
& 69.7 \\

\midrule
DP3+CORE
& 98.8
& 87.0
& 89.2
& \textbf{91.4}
& 32.0
& \textbf{88.8}
& \textbf{90.6}
& 80.6
& \textbf{92.6}
& \textbf{72.0}
& \textbf{82.3} \\

MP1+CORE
& \textbf{100.0} 
& \textbf{90.4} 
& \textbf{91.6} 
& 87.6 
& \textbf{33.6} 
& 86.8 
& 90.0 
& \textbf{89.0} 
& 55.8 
& 69.6 
& 79.4 \\
\bottomrule
\end{tabular}%
}
\caption{Performance of different methods on the RoboTwin 2.0 benchmark.}
\label{tab:task_success_transposed}
\end{table*}

\section{Experiments}
To evaluate the effectiveness of the proposed framework, CORE, we conduct comprehensive experiments comparing it against SOTA methods on two simulation benchmarks: Meta-World \cite{yu2020meta} and RoboTwin 2.0 \cite{chen2025robotwin}. Furthermore, we validate its real-world applicability on a physical robotic arm and perform extensive ablation studies to analyze the impact of key hyperparameters.

\subsection{Experimental Setup}

We evaluate CORE in a multi-task setting on Meta-World~\cite{yu2020meta} and RoboTwin 2.0~\cite{chen2025robotwin}, using success rate as the primary metric. Our Meta-World evaluation covers 39 tasks, grouped into 21 Easy, 9 Medium, 5 Hard, and 4 Very Hard tasks. RoboTwin 2.0 contains 10 representative tasks covering object placement, handover, and stacking.

We compare against VIP~\cite{ma2022vip}, DP~\cite{chi2025diffusion}, DP3~\cite{Ze2024DP3}, FlowPolicy~\cite{zhang2025flowpolicy}, and MP1~\cite{sheng2026mp1}. CORE is applied to DP3 and MP1, yielding DP3+CORE and MP1+CORE. We also evaluate text-conditioned counterparts, DP3+TEXT and MP1+TEXT, which use the same policy backbones and conditioning interface as CORE, but replace $z_{\mathrm{goal}}$ with a frozen text embedding of the task instruction. For 3D policy backbones, all methods use the same point-cloud observations. VIP and DP are included as image-based reference baselines following their original observation settings.

On Meta-World, we use 10 robot demonstrations and 50 action-free visual demonstrations, and report the mean and standard deviation over three random seeds. On RoboTwin 2.0, we use 50 robot demonstrations and 50 action-free visual demonstrations. Following the evaluation protocols for the baselines we compare, we report results from a single training seed. Full hyperparameters, training schedules, evaluation frequency, and hardware details are provided in the supplementary material.

\subsection{Simulation Evaluation}
\textbf{Results on Meta-World.}
Table~\ref{results} shows that CORE improves both policy backbones. With MP1, CORE increases the average success rate from 81.7\% to 85.6\% (+3.9 points); with DP3, CORE improves the average success rate from 81.5\% to 84.0\% (+2.5 points). In comparison, text-conditioned variants provide only limited gains, with DP3+TEXT and MP1+TEXT reaching 81.9\% and 81.4\%, respectively. The improvement is more pronounced on difficult tasks. For example, MP1+CORE improves Hard tasks from 55.1\% to 64.0\% and Very Hard tasks from 75.5\% to 81.4\%. These results suggest that visual outcome prototypes extracted from successful demonstrations provide more concrete and actionable constraints than high-level language descriptions, especially for tasks requiring precise terminal-state alignment. Additionally, VIP, which relies on terminal states for implicit reward learning, achieves only a 29.3\% average success rate, highlighting the advantage of CORE's explicit goal conditions.

\begin{figure}[t]
    \centering
    \includegraphics[width=1.0\linewidth]{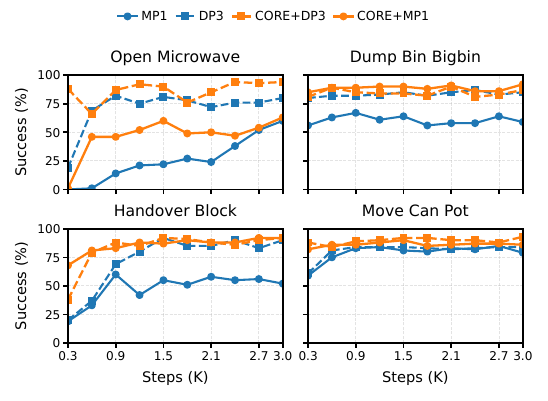}
    \caption{
    Training curves on representative RoboTwin 2.0 tasks. CORE-based policy yields higher, more stable success rates than vanilla DP3 and MP1, with varying margins.
}
    \label{fig:train_step}
\end{figure}

\textbf{Results on RoboTwin 2.0.}
Table~\ref{tab:task_success_transposed} further evaluates CORE on 10 RoboTwin 2.0 tasks. DP3+CORE achieves an average success rate of 82.3\%, improving over DP3 by +4.8 points and over DP3+TEXT by +4.9 points. MP1+CORE reaches 79.4\%, outperforming MP1 by +11.1 points and MP1+TEXT by +9.7 points. In contrast, VIP achieves an average of only 42.9\%, further showing that implicit visual rewards struggle in contact-rich settings. At the task level, at least one CORE-integrated policy achieves the best result on each evaluated task. Fig.~\ref{fig:train_step} further visualizes the training dynamics on four representative RoboTwin tasks. 
The CORE variants tend to achieve higher or more stable success rates as training progresses, although the magnitude of improvement varies across tasks and backbones. 
These curves are consistent with the task-level results in Table~\ref{tab:task_success_transposed}, suggesting that visual outcome prototypes provide a useful optimization signal rather than merely improving the final reported average.

\textbf{Summary and Analysis.}
Overall, the results across both benchmarks show that CORE improves manipulation policies by providing outcome-level visual guidance rather than adding another input modality. Language conditions can describe task intent, but they often fail to specify the exact object poses, spatial relations, and contact constraints required for successful manipulation. CORE instead learns such constraints from successful terminal states in visual demonstrations and converts them into reusable visual goal prototypes. This explains why CORE brings modest gains on easier tasks where baseline policies perform well, but yields larger improvements on tasks that require precise final-state control. These findings support the key idea of CORE: action-free visual demonstrations can guide robot policies effectively when exploited through shared outcome regularities.

\begin{table}[t]
\centering
\caption{Component ablation on Meta-World Hard and Very Hard tasks. 
All variants use MP1 as the policy backbone and share the same goal prototype construction.}
\label{tab:ablation_component}
\small
\setlength{\tabcolsep}{4pt}
\begin{tabular}{lcccc}
\toprule
Variant & Hard & Very Hard & Avg. & Drop \\
\midrule
MP1 & 55.1 & 75.5 & 64.2 & 7.6 \\
MP1+CORE & \textbf{64.0} & \textbf{81.4} & \textbf{71.7} & -- \\
\midrule
w/o $L_{\mathrm{ttg}}$ & 57.3 & 76.7 & 65.9 & 5.8 \\
w/o $L_{\mathrm{term}}$ & 57.5 & 77.1 & 66.2 & 5.5 \\
w/o $L_{\mathrm{ttg}}, L_{\mathrm{term}}$ & 54.8 & 74.2 & 63.4 & 8.3 \\
w/o $z_t$ & 59.4 & 78.8 & 68.0 & 3.7 \\
w/o $z_{goal}$ & 56.3 & 77.7 & 65.8 & 5.9\\
Random $z_{goal}$ & 41.8 & 53.0 & 46.8 & 24.9\\
\bottomrule
\end{tabular}
\end{table}

\begin{table}[t]
\centering
\caption{Ablation of goal prototype construction on Meta-World Hard and Very Hard tasks. All variants share the MP1 backbone and training settings, varying only $z_{\mathrm{goal}}$.}
\label{tab:ablation_zgoal}
\small
\setlength{\tabcolsep}{4pt}
\begin{tabular}{lccc}
\toprule
$z_{\mathrm{goal}}$ construction & Hard & Very Hard & Avg. \\
\midrule
Random terminal embedding & 62.7 & 80.3 & 70.5 \\
Single prototype (mean) & 63.8 & 80.9 & 71.4 \\
Top-K prototype averaging & \textbf{64.0} & \textbf{81.4} & \textbf{71.7} \\
\bottomrule
\end{tabular}
\end{table}

\subsection{Ablation Study}

We conduct ablation studies to identify the sources of improvement in CORE. Unless otherwise specified, all variants follow the same robot policy training setting and evaluation protocol as the main experiments. We study four aspects of the framework: the auxiliary objectives and the current outcome embedding, the construction of the goal prototype $z_{\mathrm{goal}}$, the number of visual demonstrations without action labels, and the visual encoder used in Stage I.

\textbf{Component ablation.}
Table~\ref{tab:ablation_component} reports the component ablations on the
Hard and Very Hard Meta-World tasks. The full CORE model achieves an average
success rate of 71.7\%, compared with 64.2\% for MP1. Removing
$\mathcal{L}_{\mathrm{ttg}}$ or $\mathcal{L}_{\mathrm{term}}$ reduces the
average success rate to 65.9\% and 66.2\%, respectively, showing that both
trajectory-progress supervision and terminal discrimination contribute to
learning a useful outcome representation. Removing both losses further
degrades performance to 63.4\%, even below MP1, highlighting their importance
in structuring the terminal representation space. Removing $z_t$ lowers the
average success rate to 68.0\%, indicating that a static goal prototype alone
lacks information about the current outcome. Conversely, removing
$z_{\mathrm{goal}}$ reduces performance to 65.8\%, showing that $z_t$ alone
cannot specify the desired terminal outcome. Replacing the learned
$z_{\mathrm{goal}}$ with a random vector causes a substantially larger drop
to 46.8\%, demonstrating that the gain arises from the task-relevant outcome
structure encoded by the learned prototype rather than merely from adding an
extra conditioning vector. Overall, these results support both the
representation supervision in Stage I and the complementary roles of $z_t$
and $z_{\mathrm{goal}}$ in the Stage-III policy condition
$g_t=[z_t;z_{\mathrm{goal}}]$.

\textbf{Goal prototype construction.}
Table~\ref{tab:ablation_zgoal} compares different strategies for constructing $z_{\mathrm{goal}}$ from successful terminal embeddings. Randomly selecting one terminal embedding obtains 70.5\% average success, showing that a single successful outcome contains useful goal information, but this strategy can be sensitive to the sampled demonstration. Averaging all successful terminal embeddings improves the average success rate to 71.4\%, suggesting that aggregating multiple demonstrations helps reduce single-sample noise. However, direct averaging treats all terminal embeddings equally and may still be affected by noisy samples or less representative terminal modes. The Top-K prototype averaging achieves the best result, with 64.0\% on Hard tasks, 81.4\% on Very Hard tasks, and 71.7\% on average. This strategy first clusters successful terminal embeddings with K-Means and then averages the selected Top-K representative prototypes, which preserves dominant successful outcome modes while suppressing noisy or low-frequency clusters. Although the improvement over direct averaging is moderate, the consistent gains on Hard and Very Hard tasks indicate that filtering representative prototypes before averaging provides a more robust static goal condition for difficult manipulation tasks.

\begin{figure}[t]
    \centering
    \includegraphics[width=1.0\linewidth]{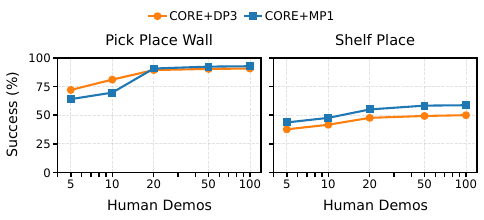}
    \caption{Effect of action-free human demonstration quantity (5 to 100) used for outcome representation and prototype learning, with policy settings fixed.}
\label{fig:ablation_demo_num}
\end{figure}

\textbf{Number of visual demonstrations.}
Fig.~\ref{fig:ablation_demo_num} studies how the number of visual demonstrations without action labels affects CORE. We vary the number of visual demonstrations from 5 to 100 while keeping the robot policy training data unchanged. On Pick Place Wall, both CORE+DP3 and CORE+MP1 improve quickly as more visual demonstrations are added, and the curves become relatively stable after about 20 demonstrations. On Shelf Place, the increase is more gradual, which suggests that additional demonstrations help cover a broader set of successful terminal configurations. These results show that CORE can extract useful outcome information from limited visual data, while more demonstrations improve the coverage and stability of the visual goal prototypes.

\begin{figure}[t]
    \centering
    \includegraphics[width=1\linewidth]{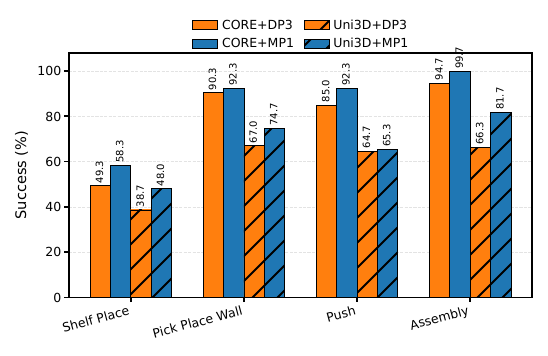}
    \caption{Ablation on the Stage-I visual encoder. 
    CORE denotes the learned terminal outcome encoder, while Uni3D denotes a frozen pretrained visual encoder used under the same goal prototype construction and policy training pipeline.}
\label{fig:encoder_ablation}
\end{figure}

\textbf{Visual encoder in Stage I.}
Fig.~\ref{fig:encoder_ablation} compares the learned terminal encoder in CORE with a frozen Uni3D encoder under the same goal prototype construction and policy training pipeline. Across Shelf Place, Pick Place Wall, Push, and Assembly, the learned CORE encoder consistently outperforms the general purpose visual encoder for both policy backbones. Averaged over the four tasks, CORE+DP3 achieves 79.8\% success, while Uni3D+DP3 obtains 59.2\%; CORE+MP1 achieves 85.7\%, while Uni3D+MP1 obtains 67.4\%. This gap indicates that the gain of CORE does not simply come from adding visual features. Instead, the encoder needs to be trained around successful outcomes so that the representation can focus on terminal structures, spatial relations, and contact patterns that matter for manipulation.

Overall, the ablation results show that the three stages of CORE play complementary roles. Stage I learns a terminal representation space with temporal progress and terminal discrimination. Stage II converts successful terminal embeddings into visual goal prototypes. Stage III conditions the policy on both the current outcome embedding and the desired goal prototype. The studies on human demonstration number and visual encoder choice further show that CORE can effectively use human data without action labels, and that learning representations around successful outcomes is more effective than directly reusing generic visual features.

\begin{figure}[t]
    \centering
    \includegraphics[width=0.6\linewidth]{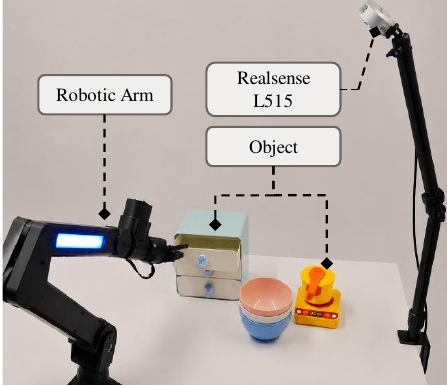}
    \caption{Real-world setup.}
    \label{fig:setup}
\end{figure}

\begin{table}[t]
\centering
\scriptsize
\setlength{\tabcolsep}{3.5pt}
\renewcommand{\arraystretch}{1.1}
\caption{Real-world manipulation success rate (\%) on five tasks.}
\label{tab:realworld}
\begin{tabular}{l c c c c c c}
\hline
\textbf{Method} & \textbf{Adj Bot} & \textbf{Cook} & \textbf{Cls Drw} & \textbf{Opn Drw} & \textbf{Stack3} & \textbf{Avg} \\
\hline
DP3        & 60 & 80 & 65 & 45 & 75 & 65.0 \\
MP1        & 70 & 75 & 70 & 55 & 75 & 69.0 \\
DP3+TEXT   & 60 & 80 & 75 & 50 & 75 & 68.0 \\
MP1+TEXT   & 65 & 80 & 70 & 50 & 70 & 67.0 \\
\hline
DP3+CORE   & 75 & 95 & 85 & 70 & 85 & 82.0 \\
MP1+CORE   & 85 & 90 & 90 & 75 & 85 & 85.0 \\
\hline
\end{tabular}
\end{table}

\begin{table}[t]
\centering
\scriptsize
\setlength{\tabcolsep}{2.5pt}
\caption{Real-world MP1 success rates (\%) with different action-free (AF) visual data sources (20 trials/task).}
\begin{tabular}{lcccccc}
\toprule
AF source & Adj. & Cook & Close & Open & Stack3 & Avg. \\
\midrule
None  & 70 & 75 & 70 & 55 & 75 & 69.0 \\
Robot & 85 & 90 & 90 & 80 & 90 & \textbf{87.0} \\
Human & 80 & 85 & 85 & 80 & 90 & 84.0 \\
Mixed & 85 & 90 & 90 & 75 & 85 & 85.0 \\
\bottomrule
\end{tabular}

\label{tab:real_action_free_source}
\end{table}

\subsection{Real-World Experiments}
\textbf{Hardware \& Task Descriptions.} We deploy our policies on an ARX R5 robotic arm. Visual observations are captured via an Intel RealSense L515 camera, which provides point cloud data of the workspace. For training, we collect 100 robot demonstrations and 50 action-free human demonstrations per task. During evaluation, we perform 20 rollouts per task. We evaluate five representative contact-rich manipulation tasks: Adjust Bottle (Adj Bot), Cook, Close Drawer (Cls Drw), Open Drawer (Opn Drw), and Stack 3 Bowls (Stack3). These tasks heavily rely on precise geometric alignment and physical interactions.

\textbf{Experimental Results.} As shown in Table~\ref{tab:realworld}, MP1+CORE and DP3+CORE achieve average success rates of 85.0\% and 82.0\%, yielding absolute improvements of 16.0\% and 17.0\% over their vanilla counterparts, respectively. Conversely, text-conditioned variants (DP3+TEXT, MP1+TEXT) provide negligible gains. These quantitative results are further corroborated by qualitative visualizations (Fig.~\ref{fig:real_qualitative}). For instance, in the Open Drawer and Close Drawer tasks, the vanilla MP1 policy often stops slightly short of the target. By leveraging Common Outcome Regularities, CORE actively guides the robot to successfully complete the tasks. Ultimately, this demonstrates that visual goal prototypes extracted by CORE offer much more precise physical constraints for real-world execution than abstract language instructions.
Furthermore, Table~\ref{tab:real_action_free_source} shows that robot, human, and mixed action-free visual data achieve 87.0\%, 84.0\%, and 85.0\% average success, respectively, compared with 69.0\% without action-free data, demonstrating that CORE can exploit visual supervision from multiple embodiments.

\section{Conclusion}
We presented CORE, an outcome-level framework for using action-free visual demonstrations in robot manipulation. CORE learns terminal outcome representations, aggregates successful terminals into visual goal prototypes, and injects these prototypes into robot policy backbones as concrete goal conditions. Extensive evaluations across simulation and real-world settings show that CORE consistently improves manipulation success rates by leveraging additional action-free visual demonstrations while keeping the amount of action-labeled robot data fixed.

\bibliography{aaai2027}


\end{document}